# Equine radiograph classification using deep convolutional neural networks

Costa da Silva, Raniere Gaia; Department of Infectious Diseases and Public Health, City University of Hong Kong, Tat Chee Avenue, Kowloon, Hong Kong SAR China

Mishra, Ambika Prasad; Department of Infectious Diseases and Public Health, City University of Hong Kong, Tat Chee Avenue, Kowloon, Hong Kong SAR China

Riggs, Christopher; Hong Kong Jockey Club, Department of Veterinary Clinical Services

Doube, Michael; Department of Infectious Diseases and Public Health, City University of Hong Kong, Tat Chee Avenue, Kowloon, Hong Kong SAR China

2022-04-25

## 1    Abstract

**Purpose**: To assess the capability of deep convolutional neural networks to classify anatomical location and projection from a series of 48 standard views of racehorse limbs.

**Materials and Methods**: 9504 equine pre-import radiographs were used to train, validate, and test six deep learning architectures available as part of the open source machine learning framework PyTorch.

**Results**: ResNet-34 achieved a top-1 accuracy of 0.8408 and the majority (88%) of misclassification was because of wrong laterality. Class activation maps indicated that joint morphology drove the model decision.

**Conclusion**: Deep convolutional neural networks are capable of classifying equine pre-import radiographs into the 48 standard views including moderate discrimination of laterality independent of side marker presence.

## 2    Introduction

Radiomics is an emerging technology with the potential to augment radiological diagnosis and research ([1],[2]). So far, most radiomic studies are based on human patients ([3]). Humans have a limited range of morphological variation, while centralisation of radiographic services in large hospitals leads to imaging datasets that are consistent and numerous enough to be used in the development of machine learning models (4). Veterinary species are typically radiographed in decentralised small practices with a relatively much lower case-load than human radiographic practices, with highly variable ranges of views and examination conditions. Within-species variations may be large such as in dogs, which have a tremendous range of sizes and morphologies. As such, obtaining sufficient numbers and consistency of images is difficult and machine learning in veterinary radiology presents a diverse set of challenges.

Thoroughbred racehorses in Hong Kong are a useful population, because they are are all managed by the Hong Kong Jockey Club in centralised facilities (around 1200 horses at any one time).  This single breed of fit, young, mostly castrated males  represents an unusually homogeneous and large study population that is under continuous veterinary supervision by in-house clinicians. All Hong Kong racehorses must undergo a pre-import veterinary exam that includes radiography of the limbs to screen for orthopaedic anomalies (5). A standard set of 48 radiographs is reviewed to assess the horse's suitability for racing prior to importation to Hong Kong. Given the high values of these animals,  financial costs of later career-disrupting injury, and their welfare, radiographs are taken to a high standard of consistency and given close scrutiny. Radiograph sets are stored in a specialized database, *Asteris Keystone*. To our knowledge there are no extant radiomic tools in use for equine radiology.

Our objective is to establish an end-to-end artificial intelligence-based diagnostic assistant for the analysis of the 48-image pre-import radiographic series used in the Hong Kong Jockey Club protocol. In this initial study we aim to determine the most promising architectures for our application.  Here, we investigate the accuracy of current deep convolutional neural networks to classify equine pre-import radiographs into the 48 standard views, including left-right and forelimb-hindlimb discrimination. Limb anatomy distal to the carpus (wrist; colloquially 'knee') and tarsus (ankle; colloquially 'hock') is nearly identical between forelimb and hindlimb with only the third digit present and vestigial second and fourth metapodials (6). The metacarpophalangeal and metatarsophalangeal joints (fetlocks) are examined in detail due to their frequent involvement in career-limiting injury (7).

# 3 Materials and methods

## 3.1 Study Design

The Hong Kong Jockey Club would retrospectively collect and provide pre-import radiographic sets to us. We would select complete sets to use and no further selection criteria would be applied. Given the nature of anonymised, retrospective studies being conducted in this project, our institutional ethics committee determined that their approval was not required.

## 3.2 Image Acquisition

The Hong Kong Jockey Club provided anonymised radiographic image sets from 212 racehorses that had been submitted for pre-import examination. Each set was made by one of 10 veterinary clinics and comprised 48 or more radiographs of the appendicular skeleton in DICOM format with acquisition modality attribute value of CR or DX (5). We extracted metadata detailing the anatomical region and projection ('view') from the DICOM headers (using the pydicom library (version 2.2.2)) or from text burned in the image pixels (manually retyping using Label Studio) and stored it in a CSV file. Burned-in text was redacted using ImageJ (version 1.53d)  by placing a rectangular ROI over the text area and filling with black (0) pixels.  View information was processed using Python (version 3.9.9) and Pandas (version 1.3.5) to standardise the view names and to verify if the set was complete. A set was complete only if it contained exactly one of all 48 specified views, leading to a perfectly balanced data set. The Python script reported 198 sets as complete and 14 sets as incomplete. The 198 complete sets were split using Python (version 3.9.9) and scikit-learn (version 1.0.2) into training (116), validation (40), and test (42) sets. Only (19.3%) of the radiographs had a side marker indicating laterality (left of right limb) (figure 3.1).

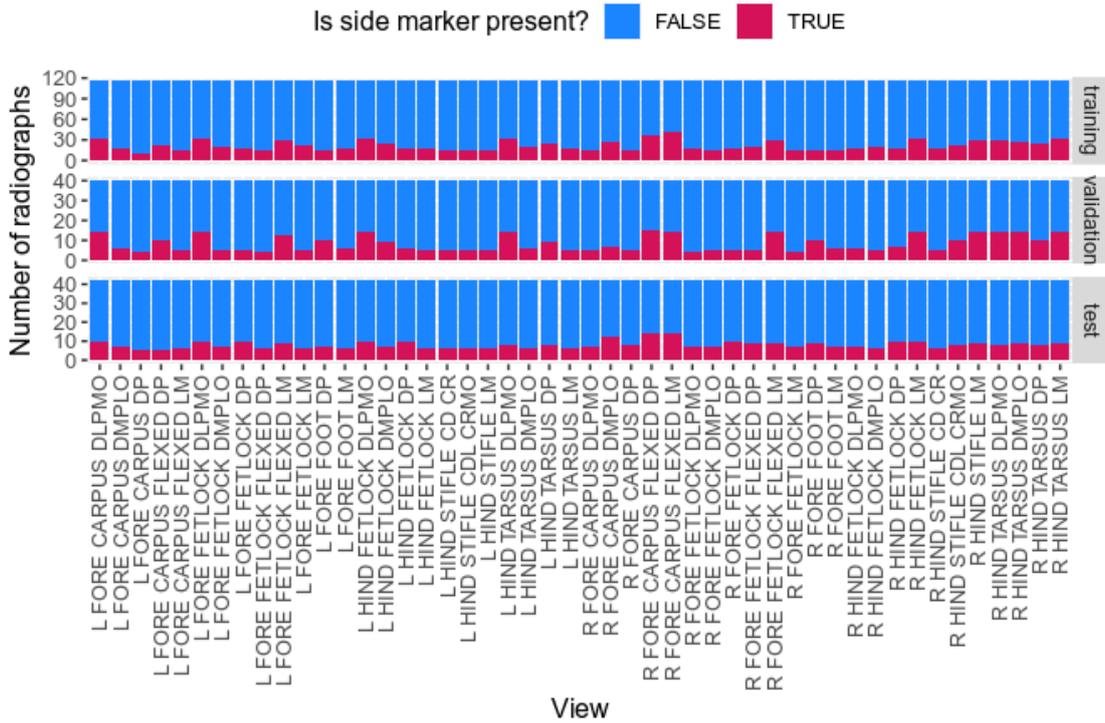

*Figure 3.1: Distribution of radiographs with side marker (left of right limb) among all views within the training, validation and test image sets.*

## 3.3   Data Preparation

The radiographs of the 198 complete sets were rotated by 90° increments and/or mirrored into the standard anatomical viewing orientation, centered on a square black (pixel value 0) canvas fit to the radiograph's long axis, downsampled to 250 × 250 pixels by nearest-neighbour interpolation from the input image,  and stored in 16-bit Tag Image File Format (TIFF) with a greyscale look-up table stretched to display 0 as black and the radiograph's maximum pixel value as white (figure 3.2; full gallery in figshare under DOI 10.6084/m9.figshare.c.5921813).

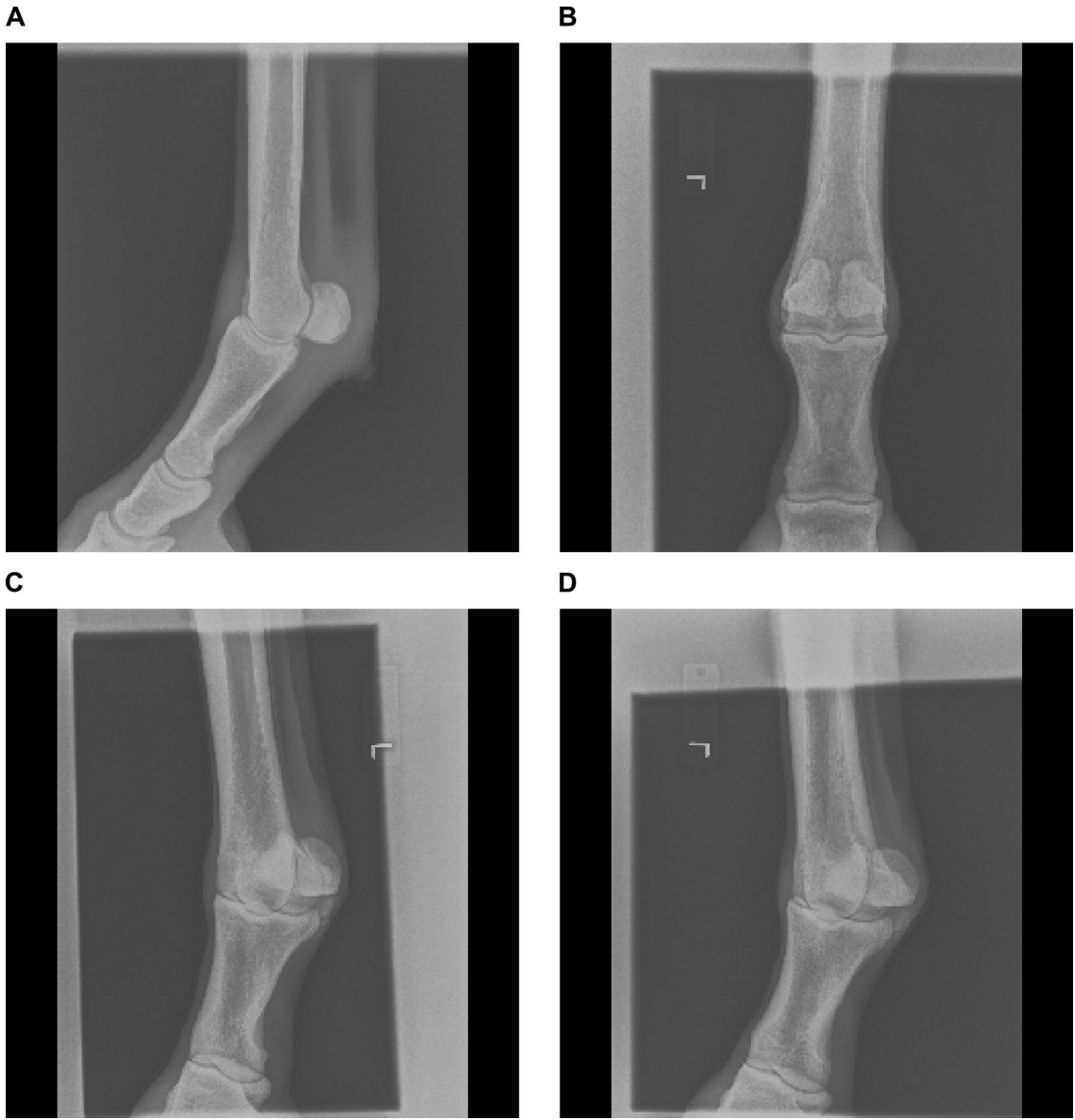

*Figure 3.2: Example of radiographs after pre-processing into 250 × 250 pixel 16-bit single channel image. (A) Left Fore Fetlock LM. (B) Left Fore Fetlock DP. (C) Left Fore Fetlock DLPMO. (D) Left Fore Fetlock DMPLO. Note the similar appearance of the two oblique views, (C, D).*

## 3.4    Deep Convolutional Neural Network Architecture

Based on the deep learning architectures comparison by (8) and (9), we adapted six deep learning architectures (table 3.1), available as part of the open source machine learning framework PyTorch (10), to output a vector with 48 values equal to the probability of the input image belonging to each of the 48 standard radiographic view. In our implementation, the 250 × 250 pixels single channel image is augmented using random zoom, random spatial crop, and random histogram shift and scaled down to 224 × 224 pixels. We trained each of the selected deep learning architectures for 128 epochs with batch size of 32, using SGDM as the optimiser. Computation was performed on a

Precision 5820 workstation (Dell Hong Kong) equipped with an Intel(R) Xeon(R) W-2123 3.60GHz CPU, 64GiB (4 × 16GiB DDR4 2666MHz) of RAM, and NVIDIA Quadro RTX 4000 GPU running Ubuntu 21.10 (Linux 5.13), Python (version 3.9.9), NumPy (version 1.21.2), PyTorch (version 1.10.2), and CUDA (version 11.3).

*Table 3.1: Deep learning models used for classification.*

| Architecture | Number of Parameters | Relative Number of Parameters |
|---|---|---|
| DenseNet-121 | 7978856 | 0.29 |
| Inception V3 | 27161264 | 1.00 |
| MobileNet V3 | 5483032 | 0.20 |
| ResNet-18 | 11689512 | 0.43 |
| ResNet-34 | 21797672 | 0.80 |
| ResNet-50 | 25557032 | 0.94 |

## 3.5   Statistical Analysis

After model training, each model was tested, and top-1 accuracy and area under the receiver operating characteristic curve (AUC) calculated using scikit-learn (version 1.0.2) with Python (version 3.9.9) and NumPy (version 1.21.2), and stored in Comet (https://www.comet.ml) for further analysis.

## 3.6   Class Activation Mapping

To gain insight into the features used by the model to determine the view classification probabilities, a class activation map for the best performing deep learning architecture was generated for each radiograph in the test set using Monai (version 0.8.0) and matplotlib (version 3.5.1) with Python (version 3.9.9), NumPy (version 1.21.2), PyTorch (version 1.10.0), and Cuda (version 11.3).

## 3.7   Model Availability

A copy of the source code of each of the six deep learning architectures is provided in a ZIP archive file (table 3.1).

# 4    Results

The best performing architecture for the classification of radiographic view was ResNet-34, which achieved a top-1 accuracy of 0.8408, with ResNet-18 slightly lower at 0.8323 (Table 4.1). Increased model size did not result in higher accuracy, with ResNet-50 achieving accuracy of only 0.8194. ResNet-34 achieved higher top-1 accuracy at earlier epochs than the other models during training and validation (figure 4.1). MobileNet V3 had a low accuracy at the begining of the training but finished with accuracy above 0.8 (figure 4.1) despite being the smallest architecture (table 3.1). ResNet-34 was retained for further analysis and the other models discarded.

*Table 4.1: Architectures, sorted by accuracy, for the classification of racehorse body part including view and laterality from radiographs and their respective metrics.*

| Architecture | Accuracy | ROC AUC |
|---|---|---|
| ResNet-34 | 0.8408 | 0.9977 |
| ResNet-18 | 0.8323 | 0.9981 |
| ResNet-50 | 0.8194 | 0.9978 |
| MobileNet V3 | 0.8095 | 0.9969 |
| DenseNet-121 | 0.7917 | 0.9971 |
| Inception V3 | 0.7371 | 0.9956 |

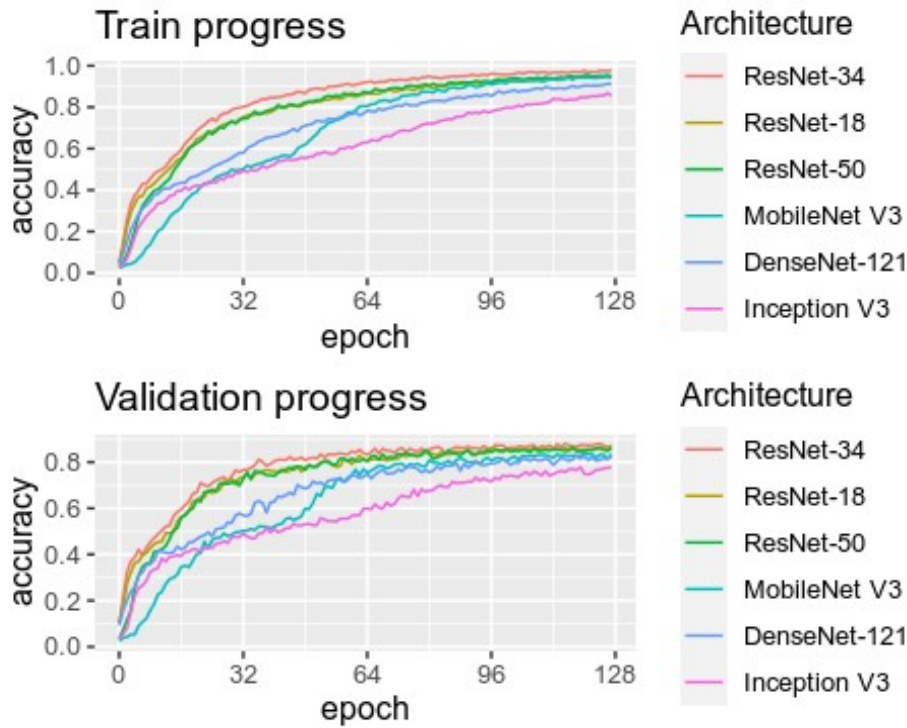

*Figure 4.1: Progress of metrics during training to classify racehorse body part including view and laterality from radiographs. An epoch is one pass over all images.*

For ResNet-34, the most common misclassification (88.8% of cases) involved assignment of a correct view to the wrong laterality (figure 4.2). Side marker was present in 19% of the radiographs during test and 17.5% of corrected classified radiographs used for test, and the side marker frequency in the two sets is significantly different. Side markers are concentrated in the misclassification group and, on average, side marker presence has a negative correlation with correct classification. Correlation between correct classification of the radiograph and side marker present in it was tested and confirmed using Pearson's Chi-squared test from R version 4.1.2 (2021-11-01): $X^2$ (1, N = 2016) = 16.3, p = 5.4e-05. Correlation between correct classification of the radiograph and side marker present varies with the view (Pearson's Chi-squared test results in supplemental material) because the number of radiographs with side marker present in the training set is different for each view (figure 3.1) and the model learnt to expect the side marker differently for each view.

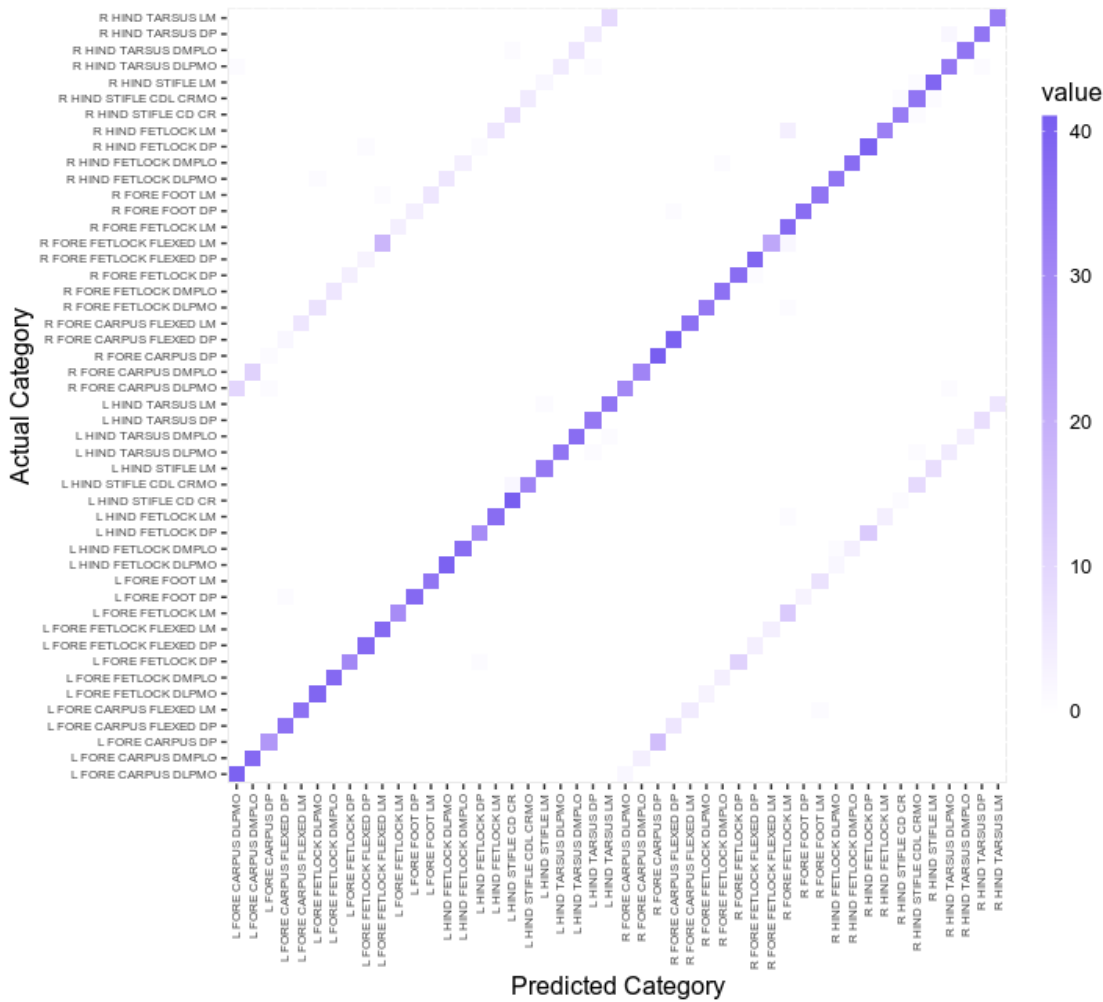

*Figure 4.2: Confusion matrix for classification of racehorse body part including view and laterality from radiographs by the best performing architecture, ResNet-34. The bright diagonal line indicates a high degree of correct classification, while the two less bright parallel diagonal lines indicate predictions where only left or right laterality was misclassified but the view and fore-hind designation was correct.*

Although Pearson's Chi-squared test indicates that side marker presence is correlated to the classification being true positive, class activation maps (11) indicated that anatomical features, in particular joint morphology, were the primary source of signal used by the model to determine view classification, and that side markers, when present, were not strongly weighted by the model (figure 4.3; full gallery in figshare under DOI 10.6084/m9.figshare.c.5921816).

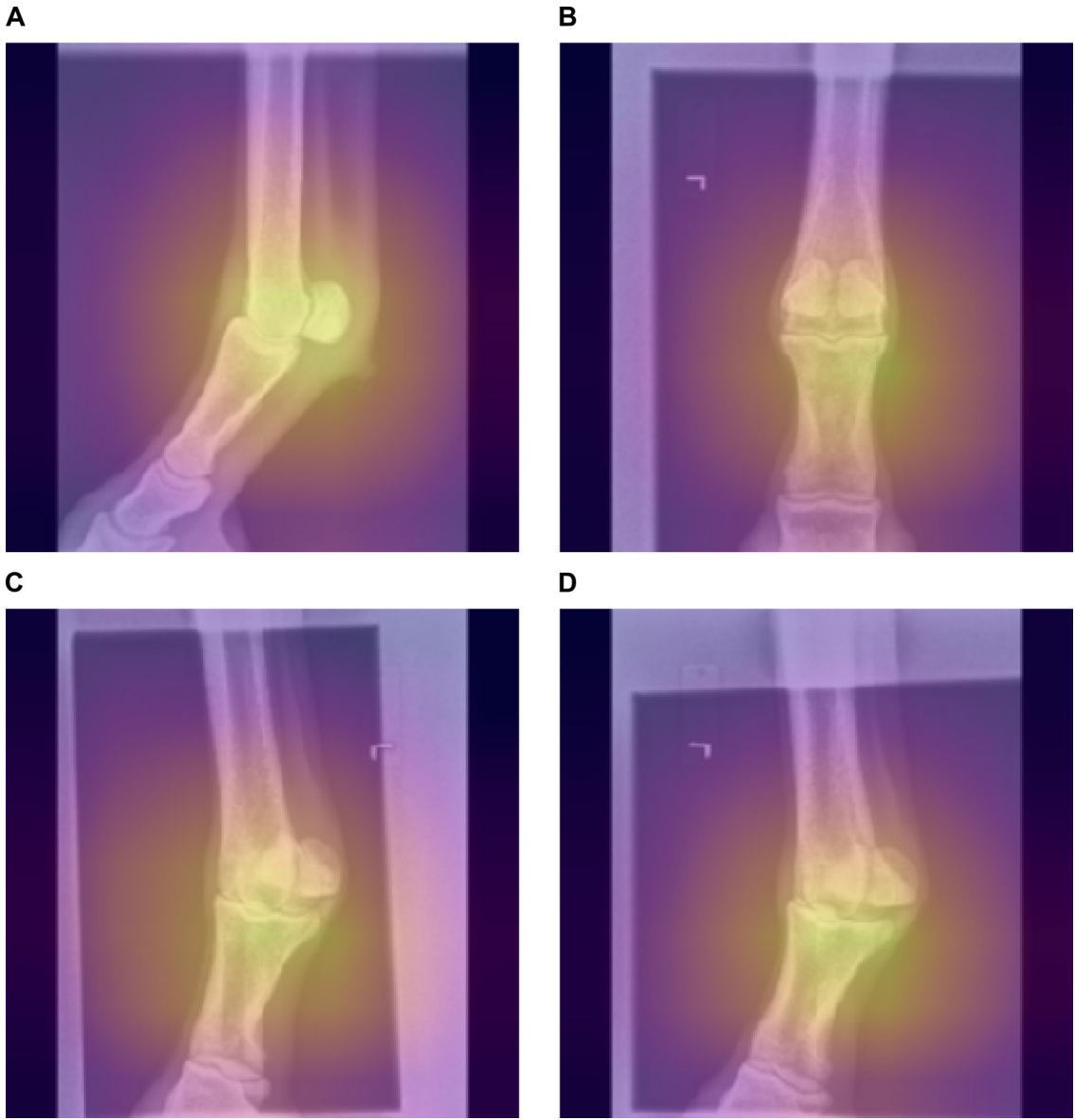

*Figure 4.3: Class activation maps applied as overlays to 224 × 224 pixel radiographs. Yellow indicates higher contribution, and purple a lower contribution, of the image region to the classification. Note that the 'L' side marker is not highlighted by the class activation map indicating that it did not contribute to the model weighting, while the metacarpophalangeal joint (fetlock) is brightly identified. (a) Left Fore Fetlock LM. (b) Left Fore Fetlock DP. (c) Left Fore Fetlock DLPMO. (d) Left Fore Fetlock DMPLO.*

Regarding redaction of radiographs to remove burned-in text, redaction was performed in 26.2% of the radiographs in the test group and in 21.9% of the correctly classified radiographs used for testing, and the redaction frequency in the two sets is significantly different. Redaction is concentrated in the misclassification group and, on average, redaction has a negative correlation with correct classification. Correlation between correct classification of the radiograph and redaction was tested and confirmed using Pearson's Chi-squared test from R version 4.1.2 (2021-11-01): $X^2$ (1, N = 2016) = 102, p = 5.7e-24. Although redaction never covered any joint morphology, we hypothesised the redaction reduced the available information in the radiograph to be used by the model which lead to a high number redacted radiographs in the misclassification group.

Performance was not evenly spread across the views, with ResNet-34 achieving > 95% best-1 accuracy (40 or more correct out of 42) when classifying left fore carpus dorsal 55º lateral to palmaromedial oblique, left hind fetlock dorsal 45º medial to pantarolateral oblique, left stifle caudocranial, right fore carpus dorsopalmar, right fore carpus flexed dorsal 60º proximal dorsodistal oblique, right hind fetlock dorsoplantar but ≤ 60% best-1 accuracy (26 or fewer correct out of 42) when classifying left fore carpus dorsopalmar, right front fetlock flexed lateromedial (details in the model report indicated in supplemental material).

If we ignore the left-right laterality classification of our model, its accuracy is 98.2% and includes the discrimination of very similar examination views, for example left fore fetlock DLPMO and left fore fetlock DMPLO (figure 3.2).

# 5   Discussion

Here, we show that deep learning architectures available in the PyTorch open source machine learning framework (10) can classify radiographic views (anatomical location and projection) of horse limbs with remarkable accuracy despite the close similarity among many of the views and heavily downsampled resolution (224 × 224 pixels) used for training, validation and testing. To our knowledge our study is the first to apply deep learning for equine limb radiographs, joining others that have investigated the application of deep learning architectures to human limb radiographs (12,13).

We defined 48 classes to fully describe the radiographs, including exam body part, exam view, and laterality, which is 6 times more classes than (12) and 3 times more than (13).  The high number of classes implies that there is a small likelihood (2.08%) of an image's view being correctly classified by chance, however, our model is correct 84.1% of the time. Accuracy is smaller than the reported by (12) (90%) and by (13) (99%, comparable to human-level error) and must be interpreted taking into consideration the differences between the data used in each of the studies, for example the presence of side markers (only 19.3% of our images contain side markers versus 99% of the radiographs used by (12) and (13)) and redaction of radiographs to remove text burned in the image pixels (28.3% of our images were redacted versus none of the radiographs used by (12) and (13)).

The most common misclassification of our model involved assignment of a correct view to the wrong laterality. (12) mentioned that the correct classification of laterality was strongly correlated with the side marker as two-thirds of the misclassification was with radiographs without a side marker and (13) mentioned that the side marker is an important classification feature based on class activation maps. Although we observed correlation between correct classification of the radiograph and side marker presence, our correlation diverges from the one reported by (12) and (13) as we observed a negative correlation and class activation maps indicate little contribution from side markers.

Random guessing of left and right should result in 50% correctly assigned (assuming no other misclassifications), however, our model detects side 84.1% correctly, implying that 34.1 times out of the remaining 50 times the model improves on chance based on anatomical information alone. It is somewhat surprising that laterality detection is better than chance, however, asymmetry of limb bone dimensions is recognized in humans (14) and *Equidae* (15), which our model appears to give weight to in its radiographic view classification. Horses worldwide are typically handled from their left ('near') side, and many racehorses do fast track work in one direction only (clockwise in Hong Kong), however, limb asymmetry also exists in horses not performing single-direction training (15,16). (17) reported significant difference in length between right and left third metacarpal bone from Thoroughbred racehorses in Victoria and South Australia. (18) reported five postmortem measurements that are larger in the left femur compared to the right femur from Thoroughbred racehorses in New South Wales, Australia. Images were supplied individually in random order and not as single-animal sets or left-right pairs, so direct within-individual comparisons were not available to the model to use in laterality discrimination. While size may be a contributing factor, yet others such as length-width aspect ratio, cortical thickness or opacity, or trabecular bone texture, might also contribute to the model weights. Individual horses commonly have bone dimensions that are asymmetric in the opposite direction to mean population-level asymmetry ([15],[16]), which may help to account for our model's imperfect accuracy in laterality classification.

Redacting radiographs to remove burned-in text in the pixels is also negatively correlated with misclassification. This is not a surprise as the redaction reduces the information available to the model. We didn't investigate this further because we expect veterinary clinics to move the patient information from the pixels to DICOM headers in the near future and redaction will not be necessary any more.

Few studies have focused on classifying radiographs based on the examination view. ([19]) used 5085 radiographs to train a model based on ResNet-152 with 96% accuracy in the task of classifying 404 human hand radiographs into posteroanterior, lateral, or oblique views, to which our model compares favourably by achieving 98.2% accuracy over 24 laterality-neutral views.

The biggest challenge to advance the application of deep learning to non-human skeletal radiographs is the lack of a large collection of radiographs publically available under an open license. In our study, we used 9,504 radiographs split into 48 different classes, which is orders of magnitude smaller than collections of human radiographs used in other studies. For example ([12]) used 256,458 radiographs of six body parts, ([13]) used 9,437 radiographs of nine body parts, ([20]) used 16,093 radiographs of five classes, and ([9]) used radiographs from MURA ([4]), a "public" dataset of musculoskeletal radiographs containing 40,561 images of seven parts of the upper limb.

In conclusion, we assessed the capacity of popular deep learning architectures to classify racehorse radiographic views, including view and laterality. The highest accuracy achieved was 0.84. Future studies will explore how to reduce the misclassification due to incorrect laterality by including additional data in the training steps, using larger images as input, and using generative adversarial networks.

# 6    Disclosures of Conflicts of Interest
- Raniere Gaia Costa da Silva: No relevant relationships.

- Ambika Prasad Mishra: No relevant relationships.

- Michael Doube: No relevant relationships.

- Christopher Riggs: No relevant relationships.

# 7    Acknowledge

We thank Dr. Anil Prabhu, Vincent Y. T. Tang, and Voss Y. T. Yu from the Hong Kong Jockey Club for provide access to the pre-import radiograph sets that were used to create the dataset used in this work.

# 8    Author Contributions

- Guarantors of integrity of entire study

  - Raniere Gaia Costa da Silva

- study concepts/study design

  - Michael Doube

  - Raniere Gaia Costa da Silva

- data acquisition

  - Raniere Gaia Costa da Silva

- data analysis/interpretation

  - Raniere Gaia Costa da Silva

- manuscript drafting or manuscript revision for important intellectual content

  - Raniere Gaia Costa da Silva

  - Michael Doube

- manuscript revision for important intellectual content

  - all authors

- approval of final version of submitted manuscript

  - all authors

- agrees to ensure any questions related to the work are appropriately resolved

  - all authors

- literature research
  - Raniere Gaia Costa da Silva
  - Michael Doube
- experimental studies
  - Raniere Gaia Costa da Silva
- statistical analysis
  - Raniere Gaia Costa da Silva
- manuscript editing
  - all authors

# 9   Funding

# 10   Ethical Statement

Permission to pre-import radiograph sets is granted by the Hong Kong Jockey Club ownership agreement.

# 11   Declaration

The paper is original, has not been submitted or published elsewhere, and has the approval of all authors.

# References

1.     Çallı E, Sogancioglu E, van Ginneken B, van Leeuwen KG, Murphy K. Deep Learning for Chest X-ray Analysis: A Survey. Medical Image Analysis. 2021;102125. doi: 10.1016/j.media.2021.102125.

2.     Mohammad-Rahimi H, Nadimi M, Rohban MH, Shamsoddin E, Lee VY, Motamedian SR. Machine learning and orthodontics, current trends and the future opportunities: A scoping review.

American Journal of Orthodontics and Dentofacial Orthopedics. 2021; doi: 10.1016/j.ajodo.2021.02.013.

3.      Awaysheh A, Wilcke J, Elvinger F, Rees L, Fan W, Zimmerman KL. Review of Medical Decision Support and Machine-Learning Methods. Vet Pathol. SAGE Publications Inc; 2019;56(4):512–525. doi: 10.1177/0300985819829524.

4.      Rajpurkar P, Irvin J, Bagul A, et al. MURA: Large Dataset for Abnormality Detection in Musculoskeletal Radiographs. 2018. http://arxiv.org/abs/1712.06957. Accessed February 5, 2021.

5.      Veterinary Regulation, Welfare and Biosecurity Policy. The Hong Kong Jockey Club Veterinary Pre-Import Examination Protocol. The Hong Kong Jockey Club; 2020. http://www.mathildetexier.com/mathildetexier.com/Home_files/HKJC%20Veterinary%20Pre-Import %20Examination%20Protocol%20September%2020.pdf. Accessed February 22, 2021.

6.      Solounias N, Danowitz M, Stachtiaris E, et al. The evolution and anatomy of the horse manus with an emphasis on digit reduction. R Soc open sci. 2018;5(1):171782. doi: 10.1098/rsos.171782.

7.      Rosanowski SM, Chang YM, Stirk AJ, Verheyen KLP. Epidemiology of race-day distal limb fracture in flat racing Thoroughbreds in Great Britain (2000-2013). Equine Vet J. 2019;51(1):83–89. doi: 10.1111/evj.12968.

8.      Canziani A, Paszke A, Culurciello E. An Analysis of Deep Neural Network Models for Practical Applications. 2017. http://arxiv.org/abs/1605.07678. Accessed June 17, 2021.

9.      Ananda A, Ngan KH, Karabağ C, Ter-Sarkisov A, Alonso E, Reyes-Aldasoro CC. Classification and Visualisation of Normal and Abnormal Radiographs; A Comparison between Eleven Convolutional Neural Network Architectures. Sensors. 2021;21(16):5381. doi: 10.3390/s21165381.

10.     Paszke A, Gross S, Massa F, et al. PyTorch: An Imperative Style, High-Performance Deep Learning Library. Vancouver, Canada; 2019. p. 12.

https://papers.nips.cc/paper/2019/hash/bdbca288fee7f92f2bfa9f7012727740-Abstract.html.
Accessed April 21, 2021.

11.     Zhou B, Khosla A, Lapedriza A, Oliva A, Torralba A. Learning Deep Features for Discriminative Localization. 2015. http://arxiv.org/abs/1512.04150. Accessed May 11, 2021.

12.     Olczak J, Fahlberg N, Maki A, et al. Artificial intelligence for analyzing orthopedic trauma radiographs. Acta Orthopaedica. Taylor & Francis; 2017;88(6):581–586. doi: 10.1080/17453674.2017.1344459.

13.     Filice RW, Frantz SK. Effectiveness of Deep Learning Algorithms to Determine Laterality in Radiographs. J Digit Imaging. 2019;32(4):656–664. doi: 10.1007/s10278-019-00226-y.

14.     Roy TA, Ruff CB, Plato CC. Hand dominance and bilateral asymmetry in the structure of the second metacarpal. American Journal of Physical Anthropology. 1994;94(2):203–211. doi: 10.1002/ajpa.1330940205.

15.     Leśniak KG, Williams JM. Relationship Between Magnitude and Direction of Asymmetries in Facial and Limb Traits in Horses and Ponies. Journal of Equine Veterinary Science. 2020;93:103195. doi: 10.1016/j.jevs.2020.103195.

16.     Leśniak K. The incidence of, and relationship between, distal limb and facial asymmetry, and performance in the event horse. Comparative Exercise Physiology. 2020;16(1):47–53. doi: 10.3920/CEP190047.

17.     Watson KM, Stitson DJ, Davies HMS. Third metacarpal bone length and skeletal asymmetry in the Thoroughbred racehorse. Equine Veterinary Journal. 2003;35(7):712–714. doi: 10.2746/042516403775696348.

18.     Pearce G, May-Davis S, Greaves D. Femoral asymmetry in the Thoroughbred racehorse. Australian Veterinary Journal. 2005;83(6):367–370. doi: 10.1111/j.1751-0813.2005.tb15636.x.

19.     Saun TJ. Automated Classification of Radiographic Positioning of Hand X-Rays Using a Deep Neural Network. Plast Surg (Oakv). SAGE Publications Inc; 2021;29(2):75–80. doi: 10.1177/2292550321997012.

20.     Vo DV, Pham HH, Nguyen HQ. Automatic Classification of Human Body Parts from X-ray Images Using Deep Convolutional Neural Networks. 2021. https://openreview.net/pdf?id=zKBgupz4o2. Accessed April 29, 2021.

# 12     Supplemental Material

## 12.1  Pre-sale Examination Set

*Table 12.1: Examination Set.*

| Anatomical Region | View (Abbreviation) | View |
|---|---|---|
| Carpus | DP | dorsopalmar |
| Carpus | DLPMO | dorsal 55º lateral to palmaromedial oblique |
| Carpus | DMPLO | dorsal 75º medial to palmarolateral oblique |
| Carpus | FLEXED LM | flexed lateromedial |
| Carpus | FLEXED DP | flexed dorsal 60º proximal dorsodistal oblique |
| Fore Fetlock | DP | dorsopalmar |
| Fore Fetlock | DLPMO | dorsal 45º llateral to palmaromedial oblique |
| Fore Fetlock | DMPLO | dorsal 45º medial to palmarolateral oblique |
| Fore Fetlock | FLEXED LM | flexed lateromedial |
| Fore Fetlock | FLEXED DP | flexed dorsal 125º distal to palmaroproximal oblique |
| Fore Fetlock | LM | lateromedial |
| Hind Fetlock | DP | dorsoplantar |
| Hind Fetlock | DLPMO | dorsal 45º lateral to pantaromedial oblique |
| Hind Fetlock | DMPLO | dorsal 45º  medial to pantarolateral oblique |

| Hind Fetlock | LM | lateromedial |
|---|---|---|
| Tarsus | DP | dorsoplantar |
| Tarsus | DLPMO | dorsal 10º lateral to pantaromedial oblique |
| Tarsus | DMPLO | dorsal 65º medial to pantarolateral oblique |
| Tarsus | LM | lateromedial |
| Stifle | LM | lateromedial |
| Stifle | CD CR | caudocranial |
| Stifle | CDL CRMO | caudolateral to craniomedial oblique |
| Fore Hoof | LM | lateromedial |
| Fore Hoof | DP | dorsal 60º proximal to palmarodistal oblique |

## 12.2  Model

*Table 12.2: Model.*

| Architecture | Source Code | Weights |
|---|---|---|
| DenseNet-121 | 10.6084/m9.figshare.19453334 | 10.6084/m9.figshare.19453307 |
| Inception V3 | 10.6084/m9.figshare.19453319 | 10.6084/m9.figshare.19453307 |
| MobileNet V3 | 10.6084/m9.figshare.19453349 | 10.6084/m9.figshare.19453307 |
| ResNet-18 | 10.6084/m9.figshare.19453274 | 10.6084/m9.figshare.19453277 |
| ResNet-34 | 10.6084/m9.figshare.19453289 | 10.6084/m9.figshare.19453292 |
| ResNet-50 | 10.6084/m9.figshare.19453304 | 10.6084/m9.figshare.19453307 |

## 12.3  Model Report

*Table 12.3: Model Report Location.*

| Architecture | Jupyter Notebook | Report |
|---|---|---|
| DenseNet-121 | 10.6084/m9.figshare.19453343 | 10.6084/m9.figshare.19453346 |
| Inception V3 | 10.6084/m9.figshare.19453328 | 10.6084/m9.figshare.19453331 |
| MobileNet V3 | 10.6084/m9.figshare.19453358 | 10.6084/m9.figshare.19453361 |

| ResNet-18 | 10.6084/m9.figshare.19453283 | 10.6084/m9.figshare.19453286 |
| ResNet-34 | 10.6084/m9.figshare.19453298 | 10.6084/m9.figshare.19453301 |
| ResNet-50 | 10.6084/m9.figshare.19453313 | 10.6084/m9.figshare.19453316 |

## 12.4  Side Marker Presence and Classification Success

Our null hypothesis is that side marker presence and classification success are dependent in the set of radiographs with a single label when the classification is provided by ResNet-34. The null hypothesis was tested using Pearson's chi-squared test implemented in R version 4.1.2 (2021-11-01) (table 12.4).

*Table 12.4: Pearson's chi-squared test of side marker presence and classification success for each label.*

| Label | With side marker (%) | Correctly classified (%) | p-value |
| --- | --- | --- | --- |
| L FORE CARPUS DLPMO | 23.8095 | 95.2381 | 0.0095 |
| L FORE CARPUS DMPLO | 16.6667 | 90.4762 | 0.6382 |
| L FORE CARPUS DP | 11.9048 | 61.9048 | 0.9256 |
| L FORE CARPUS FLEXED DP | 11.9048 | 85.7143 | 0.0800 |
| L FORE CARPUS FLEXED LM | 14.2857 | 85.7143 | 0.2801 |
| L FORE FETLOCK DLPMO | 23.8095 | 92.8571 | 0.6877 |
| L FORE FETLOCK DMPLO | 16.6667 | 90.4762 | 0.3471 |
| L FORE FETLOCK DP | 23.8095 | 71.4286 | 0.3594 |
| L FORE FETLOCK FLEXED DP | 14.2857 | 90.4762 | 0.5197 |
| L FORE FETLOCK FLEXED LM | 21.4286 | 90.4762 | 0.8548 |
| L FORE FETLOCK LM | 14.2857 | 69.0476 | 0.2757 |
| L FORE FOOT DP | 16.6667 | 90.4762 | 0.6382 |
| L FORE FOOT LM | 14.2857 | 83.3333 | 1.0000 |
| L HIND FETLOCK DLPMO | 23.8095 | 95.2381 | 0.3729 |
| L HIND FETLOCK DMPLO | 16.6667 | 88.0952 | 0.1358 |
| L HIND FETLOCK DP | 23.8095 | 69.0476 | 0.0228 |
| L HIND FETLOCK LM | 14.2857 | 88.0952 | 0.0019 |
| L HIND STIFLE CD CR | 14.2857 | 97.6190 | 0.6795 |
| L HIND STIFLE CDL CRMO | 14.2857 | 73.8095 | 0.6673 |
| L HIND STIFLE LM | 14.2857 | 80.9524 | 0.3358 |
| L HIND TARSUS DLPMO | 19.0476 | 83.3333 | 0.0789 |
| L HIND TARSUS DMPLO | 14.2857 | 88.0952 | 0.6972 |
| L HIND TARSUS DP | 19.0476 | 80.9524 | 0.6337 |
| L HIND TARSUS LM | 14.2857 | 83.3333 | 1.0000 |
| R FORE CARPUS DLPMO | 16.6667 | 71.4286 | 0.3594 |
| R FORE CARPUS DMPLO | 28.5714 | 73.8095 | 0.5055 |
| R FORE CARPUS DP | 19.0476 | 97.6190 | 0.6235 |
| R FORE CARPUS FLEXED DP | 33.3333 | 95.2381 | 0.3055 |
| R FORE CARPUS FLEXED LM | 33.3333 | 85.7143 | 1.0000 |
| R FORE FETLOCK DLPMO | 16.6667 | 80.9524 | 0.7252 |
| R FORE FETLOCK DMPLO | 16.6667 | 85.7143 | 0.2367 |
| R FORE FETLOCK DP | 23.8095 | 88.0952 | 0.3651 |
| R FORE FETLOCK FLEXED DP | 21.4286 | 92.8571 | 0.3479 |
| R FORE FETLOCK FLEXED LM | 21.4286 | 52.3810 | 0.0052 |
| R FORE FETLOCK LM | 16.6667 | 90.4762 | 0.6382 |
| R FORE FOOT DP | 21.4286 | 88.0952 | 0.9339 |
| R FORE FOOT LM | 16.6667 | 83.3333 | 0.1949 |
| R HIND FETLOCK DLPMO | 16.6667 | 83.3333 | 0.0417 |
| R HIND FETLOCK DMPLO | 14.2857 | 88.0952 | 0.6972 |
| R HIND FETLOCK DP | 23.8095 | 95.2381 | 0.4179 |
| R HIND FETLOCK LM | 23.8095 | 76.1905 | 0.7459 |
| R HIND STIFLE CD CR | 14.2857 | 78.5714 | 0.7588 |
| R HIND STIFLE CDL CRMO | 19.0476 | 83.3333 | 0.4821 |
| R HIND STIFLE LM | 21.4286 | 92.8571 | 0.6020 |
| R HIND TARSUS DLPMO | 19.0476 | 80.9524 | 0.0132 |
| R HIND TARSUS DMPLO | 21.4286 | 83.3333 | 0.1301 |
| R HIND TARSUS DP | 19.0476 | 83.3333 | 0.7252 |
| R HIND TARSUS LM | 21.4286 | 78.5714 | 0.3261 |

# Session information

Last, let's reports the version numbers of R and all the packages used.

```
sessionInfo()
## R version 4.1.2 (2021-11-01)
## Platform: x86_64-pc-linux-gnu (64-bit)
## Running under: Ubuntu 21.10
##
## Matrix products: default
## BLAS:   /usr/lib/x86_64-linux-gnu/blas/libblas.so.3.9.0
## LAPACK: /usr/lib/x86_64-linux-gnu/lapack/liblapack.so.3.9.0
##
## locale:
##  [1] LC_CTYPE=en_GB.UTF-8       LC_NUMERIC=C
##  [3] LC_TIME=en_GB.UTF-8        LC_COLLATE=en_GB.UTF-8
##  [5] LC_MONETARY=en_GB.UTF-8    LC_MESSAGES=en_GB.UTF-8
##  [7] LC_PAPER=en_GB.UTF-8       LC_NAME=C
##  [9] LC_ADDRESS=C               LC_TELEPHONE=C
## [11] LC_MEASUREMENT=en_GB.UTF-8 LC_IDENTIFICATION=C
##
## attached base packages:
## [1] stats     graphics  grDevices utils     datasets  methods   base
```

```
## 
## other attached packages:
##  [1] koRpus.lang.en_0.1-4 koRpus_0.13-8        sylly_0.1-6
##  [4] httr_1.4.2           jsonlite_1.8.0       reshape_0.8.8
##  [7] patchwork_1.1.1      forcats_0.5.1        stringr_1.4.0
## [10] dplyr_1.0.8          purrr_0.3.4          readr_2.1.0
## [13] tidyr_1.1.4          tibble_3.1.6         ggplot2_3.3.5
## [16] tidyverse_1.3.1      configr_0.3.5
## 
## loaded via a namespace (and not attached):
##  [1] bit64_4.0.5          vroom_1.5.6
##  [3] modelr_0.1.8         assertthat_0.2.1
##  [5] highr_0.9            cellranger_1.1.0
##  [7] yaml_2.3.5           pillar_1.7.0
##  [9] backports_1.3.0      lattice_0.20-45
## [11] glue_1.6.2           digest_0.6.29
## [13] rvest_1.0.2          colorspace_2.0-3
## [15] htmltools_0.5.2      Matrix_1.4-0
## [17] plyr_1.8.6           pkgconfig_2.0.3
## [19] broom_0.7.10         haven_2.4.3
## [21] bookdown_0.24        scales_1.1.1
## [23] tzdb_0.2.0           generics_0.1.1
## [25] farver_2.1.0         ellipsis_0.3.2
## [27] withr_2.5.0          RcppTOML_0.1.7
## [29] cli_3.2.0            magrittr_2.0.2
## [31] crayon_1.5.0         readxl_1.3.1
## [33] evaluate_0.14        fs_1.5.2
## [35] fansi_1.0.2          xml2_1.3.2
## [37] tools_4.1.2          data.table_1.14.2
## [39] hms_1.1.1            lifecycle_1.0.1
## [41] munsell_0.5.0        reprex_2.0.1
## [43] compiler_4.1.2       rlang_1.0.2
## [45] grid_4.1.2           rstudioapi_0.13
## [47] labeling_0.4.2       rmarkdown_2.11
## [49] gtable_0.3.0         DBI_1.1.1
## [51] R6_2.5.1             ini_0.3.1
## [53] sylly.en_0.1-3       lubridate_1.8.0
## [55] knitr_1.37           fastmap_1.1.0
## [57] bit_4.0.4            utf8_1.2.2
## [59] stringi_1.7.6        parallel_4.1.2
## [61] Rcpp_1.0.8.3         png_0.1-7
## [63] vctrs_0.3.8          wordcountaddin_0.3.0.9000
## [65] dbplyr_2.1.1         tidyselect_1.1.1
## [67] xfun_0.29
```